\documentclass[lettersize,journal]{IEEEtran}
\usepackage{amsmath,amsfonts}
\usepackage{algorithmic}
\usepackage{algorithm}
\usepackage{array}
\usepackage[caption=false,font=normalsize,labelfont=sf,textfont=sf]{subfig}
\usepackage{textcomp}
\usepackage{stfloats}
\usepackage{url}
\usepackage{verbatim}
\usepackage{graphicx}
\usepackage{cite}
\usepackage{color}
\usepackage{multirow}
\usepackage{booktabs}
\usepackage{amssymb}
\usepackage{bbding}
\usepackage[pagebackref,breaklinks,colorlinks,bookmarks=false]{hyperref}
\usepackage{float}
\usepackage{soul}
\usepackage{color, xcolor} 

\newcommand{\mc}[1]{\mathcal{#1}}

\soulregister{\cite}7 
\soulregister{\citep}7 
\soulregister{\citet}7 
\soulregister{\ref}7 
\soulregister{\pageref}7 

\def\eg{{\em e.g.}}
\def\ie{{\em i.e.}}
\def\etal{{\em et al. }}
\sethlcolor{yellow}

\def\blue#1{#1}

\begin{document}
\title{FakeTracer: Catching Face-swap DeepFakes via
Implanting Traces in Training}
\author{Pu Sun, Honggang Qi,~\IEEEmembership{Member,~IEEE}, Yuezun Li and Siwei Lyu,~\IEEEmembership{Fellow,~IEEE}
\thanks{Pu Sun and Honggang Qi are with the University of Chinese Academy of Sciences, China. e-mail: (sunpu21@mails.ucas.ac.cn;hgqi@ucas.ac.cn).}
\thanks{Yuezun Li is with the College of Computer Science and
Technology, Ocean University of China, China. e-mail: (liyuezun@ouc.edu.cn). }
\thanks{Siwei Lyu is with University at Buffalo, SUNY, USA. Email: (siweilyu@bufflao.edu).}
\thanks{Pu Sun and Honggang Qi contribute equally. Yuezun Li is {\em corresponding author}.}}

\markboth{Journal of \LaTeX\ Class Files,~Vol.~14, No.~8, August~2021}%
{Shell \MakeLowercase{\textit{et al.}}: A Sample Article Using IEEEtran.cls for IEEE Journals}


\maketitle

\begin{abstract}
Face-swap DeepFake is an emerging AI-based face forgery technique that can replace the original face in a video with a generated face of the target identity while retaining consistent facial attributes such as expression and orientation. Due to the high privacy of faces, the misuse of this technique can raise severe social concerns, drawing tremendous attention to defend against DeepFakes recently. In this paper, we describe a new proactive defense method called FakeTracer to expose face-swap DeepFakes via implanting traces in training. Compared to general face-synthesis DeepFake, the face-swap DeepFake is more complex as it involves identity change, is subjected to the encoding-decoding process, and is trained unsupervised, increasing the difficulty of implanting traces into the training phase. To effectively defend against face-swap DeepFake, we design two types of traces, sustainable trace (STrace) and erasable trace (ETrace), to be added to training faces. During the training, these manipulated faces affect the learning of the face-swap DeepFake model, enabling it to generate faces that only contain sustainable traces. In light of these two traces, our method can effectively expose DeepFakes by identifying them. Extensive experiments corroborate the efficacy of our method on defending against face-swap DeepFake.
\end{abstract}

\vspace{-0.6cm}

\section{Introduction}

\begin{figure*}[!t]
\centering
\includegraphics[width=\linewidth]{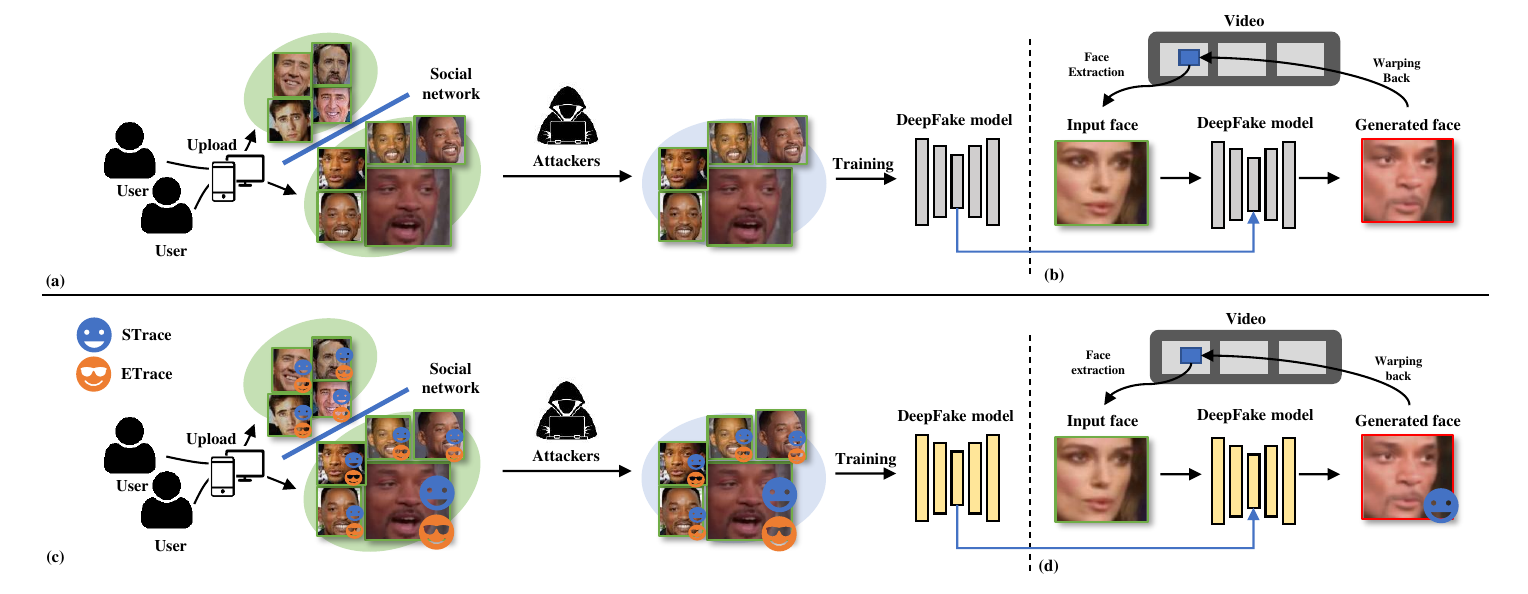}
\caption{\small (a) and (b) are the regular training phase and testing (generation) phase of the DeepFake model, while (c) and (d) are the training phase and testing (generation) phase of the DeepFake model using our method. It can be seen that our method can intervene in the DeepFake model by implanting two kinds of traces into the users' face images.}
\vspace{-0.5cm}
\label{fig:overview}
\end{figure*}

\IEEEPARstart{T}{he} significant advances in Deep Neural Networks (DNNs) have greatly improved the performance of the generative models \cite{goodfellow2014generative,kingma2014auto}, leading to a surge of AI-based face forgery techniques, known as {\em DeepFake} \cite{nguyen2019deep}. DeepFake can manipulate the attributes of original faces such as the identity \cite{liu2023deepfacelab}, facial expression and lip movements \cite{gupta2023towards} with high realism using less manual effort, causing severe concerns in society, \eg, the dependability of face recognition systems, the integrity of information on the Internet \cite{akhtar2023deepfakes}, psychological security, and political stability \cite{pantserev2020malicious} , etc. 

In recent years, lots of countermeasures have emerged to combat DeepFake, \eg, \cite{zhao2021multi, li2020face, qian2020thinking, luo2021generalizing, li2021frequency, qiao2018statistical, chen2020serial, asnani2021reverse, gu2022exploiting, wu2023interactive, guo2023rethinking, yang2021faceguard, wang2021faketagger, yu2021artificial, asnani2022proactive, yeh2020disrupting, ruiz2020disrupting, segalis2020ogan}. In particular, one typical solution for defense is DeepFake detection, which aims to identify forgery faces based on artifacts ranging from spatial to temporal domain \cite{zhao2021multi, li2020face, qian2020thinking, luo2021generalizing, li2021frequency}. However, even though they can effectively distinguish real and fake faces, they can not intervene in the generation of DeepFake faces, thus failing to prevent the creation and spread of DeepFake videos. As shown in Fig.\ref{fig:overview}(a,b), the adversary could easily access the target identity's face images from social networks and utilize them to train a DeepFake model.  

Proactive defense is a new category of countermeasures that has attracted much attention recently \cite{xue2022use, segalis2020ogan,yeh2020disrupting,ruiz2020disrupting,wang2021faketagger,yang2021faceguard,yu2021artificial,asnani2022proactive}. Compared to DeepFake detection, proactive DeepFake defense goes one step ahead to intentionally manipulate the training or testing process of the DeepFake model, in order to control the generation of DeepFake faces, such as disrupting the visual quality of generated faces using adversarial perturbations \cite{yeh2020disrupting, ruiz2020disrupting} or inserting specific watermarks into generated faces \cite{yu2021artificial}. Since we know the detail of the intervention process, the generated faces can be more easily identified.

In this paper, we focus our scope on the face-swap DeepFake, a prevalent forgery technique that can replace the original faces with newly generated faces imperceptibly (See Section \ref{sec:fswp}) and describe a new proactive method called {\em FakeTracer} to defend against face-swap DeepFakes. As shown in Fig. \ref{fig:overview}(c,d), the goal of our method is to implant specific traces into the face images of users intentionally, and these traces can significantly affect the training of DeepFake model, enabling the model to generate target faces with the desired pattern of traces. We then develop effective interpreters to detect these traces. 
Note that the existing methods either only disturbs the testing (generation) phase~\footnote{The generation phase is equivalent to the testing phase in this paper for better understanding.} \cite{asnani2022proactive, yeh2020disrupting, ruiz2020disrupting, segalis2020ogan} or focus on defending against the face-synthesis DeepFake, a technique that creates fake faces from random noise by GAN models \cite{yu2021artificial} (See Section \ref{sec:fsyn}). {To disturb the testing phase, the input face images must be processed by adding specific perturbations, which limits the practicality of defense methods, and the DeepFake model still has the potential for a generation.} In contrast, our method is devoted to implanting traces into the training of face-swap DeepFake models. {In this way, the generation ability of the DeepFake model can be affected, and the input face images in the testing phase are not required to be processed}. Unlike face-synthesis DeepFake, the face-swap DeepFake models are usually auto-encoder architectures trained in an unsupervised way, involving the identity switch and encoding-decoding process. Despite some methods of defending against 
 face-synthesis DeepFake considers manipulating the training phase, the discrepancy between face-synthesis DeepFake and face-swap DeepFake naturally results in a significant difference in defense workflows (See Section \ref{sec:challenges}).

To defend against face-swap DeepFake, we design two types of traces: sustainable trace (STrace) and erasable trace (ETrace). \blue{The STrace is an imperceptible perturbation that can be retained by the DeepFake model in the training phase, while the ETrace is a fragile perturbation that can be easily wiped off by the DeepFake encoding process. Then we can identify the authenticity of faces by detecting these two traces. The faces are
fake if STrace is detected, while the faces are real if no traces are detected, or if both STrace and ETrace are detected (See Section \ref{sec:workflow} for details).} To achieve this goal, we develop \blue{an STrace generator} to create STrace and create an identifier to recover the binary sequence by giving the STrace and DeepFake face. The successful recovery of the binary sequence represents the DeepFake face containing the STrace. Since the detail of the face-swap DeepFake model may not be accessible in reality, we design an auto-encoder architecture to simulate the encoding-decoding process by performing face reconstruction. Once the users add these two kinds of traces to the face images before uploading them to the social networks, the DeepFake model trained using these traced face images can generate fake faces with sustainable traces but no erasable traces (See Fig.~\ref{fig:method}). Then the DeepFake faces can be exposed by deciding whether the face contains sustainable and erasable traces: the face is fake if only sustainable traces are detected, and the face is real if no traces are detected, or both sustainable and erasable traces are detected. 

Extensive experiments conducted on the Celeb-DF dataset \cite{li2020celeb} demonstrate our method's efficacy. We thoroughly study our method under various settings, considering the capacity confronting different DeepFake model architectures, strategies of implanting traces, robustness against many post-processing, etc.

The contributions of this work can be summarized as follows.
\begin{enumerate}
    \item We describe a new proactive defense method (FakeTracer) to combat the generation of face-swap DeepFake, by only implanting two kinds of traces, sustainable traces and erasable trace, into the training faces for the face-swap DeepFake model, {without modifying the model architecture, training and testing process, or other parameters.}
    \item We comprehensively summarize the difference between face-synthesis DeepFake and face-swap DeepFake, and provide insights for corresponding proactive defense.
    \item We thoroughly study the effect of our method on proactively defending against face-swap DeepFake, including the effect of different DeepFake model architectures, contamination fractions, human study on the visual quality of generated faces, strategies of generating sustainable traces, and robustness against post-processing.
\end{enumerate}
This paper extends our preliminary conference paper \cite{sun2022faketracer} in several aspects: (1) We elaborate on the difference between face-synthesis DeepFake and face-swap DeepFake in the aspects of training, testing, and architectures, as well as the challenges of defending against face-swap DeepFake in the training phase; (2) We conduct more comprehensive evaluations on defense concerning various thresholds for bit accuracy, and perform more study on various settings, including the effect of using different DeepFake model architectures, the effect of using different strategies of \blue{STrace generator}, the robustness performance against many post-processing operations, and the human study on the visual quality of generated faces; (3) More tentative experiments are conducted including the efforts on implanting traces in the local region and the performance on the faces with extreme angles.

The rest of the paper is organized as follows. Section \ref{sec:bg} introduces the backgrounds of face-swap and face-synthesis DeepFake, and reviews the related works on defending against DeepFakes. Section \ref{sec:problem} describes the problem settings, including formulation, defense workflow, and the challenges of our method. Section \ref{sec:valid} describes a preliminary validation of the feasibility of our method. Section \ref{sec:strace} and \ref{sec:etrace} elaborate on the details of creating STrace and ETrace, and Section \ref{sec:exp} shows extensive experiments.

\section{Backgrounds and Related Works}
\vspace{-0.1cm}
\label{sec:bg}

\subsection{Face-swap DeepFake}
\label{sec:fswp}
\begin{figure}[!t]
\centering
\includegraphics[width=0.8\linewidth]{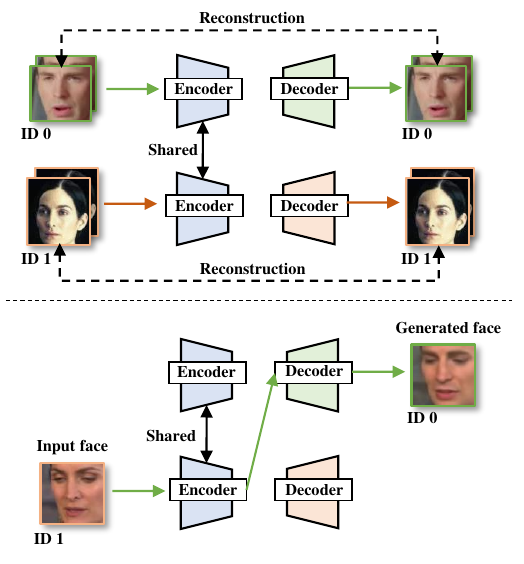}
\caption{\small The training phase (top) and testing phase (bottom) of the face-swap DeepFake model. In this illustration, ID 0 is the target identity, and ID 1 is the source identity.}
\vspace{-0.5cm}
\label{fig:face-swap}
\end{figure}

\vspace{-0.1cm}
DeepFake stands for Deep Learning and Fake Face, which stems from a Reddit account in 2017 \cite{nguyen2019deep}. Originally, DeepFake was a face-swapping technique that could generate a face of the target identity with high realism and replace it with the face of the source identity in a video while retaining consistent facial attributions such as facial expressions and orientations. The basic architecture in DeepFake is an auto-encoder \cite{kingma2014auto}. The encoder can remove the identity-related attributes but retain other attributes, such as facial expressions and orientations. The decoder can recover the appearance of the target identity on the retained facial attributes. The training of the face-swap model is usually in an unsupervised way \cite{liu2017unsupervised}. Given two independent face sets of source and target identity, one encoder for both identities is utilized, and two encoders corresponding to each identity are created. For one iteration in training, a face of a particular set is forwarded into the encoder to create a latent representation. This representation is sent into the corresponding decoder for face reconstruction. For the next iteration, a face from another set is forwarded into the encoder and another decoder for face reconstruction. These two identities are trained alternately until the generation quality is satisfied. In the testing phase (generation phase), we can select the encoder and a desired decoder as the face-swap DeepFake model. This training pipeline is simple and effective, which is still widely used in current open-source face-swap tools, \eg, FaceSwap\footnote{\url{https://faceswap.dev/}}, DeepFaceLab\footnote{\url{https://github.com/iperov/DeepFaceLab}}. These tools attract tremendous attention due to their low-cost deployment and friendly usage (each has more than $30,000$ stars on GitHub). Note that this technique can impersonate the behavior of the target identity. Thus the abuse of it can cause severe threats, such as making revengeful pornographic videos or forging unreal comments of public figures \cite{nguyen2019deep}. Fig.~\ref{fig:face-swap} shows the detail of training and testing of the face-swap DeepFake model.

\vspace{-0.3cm}
\subsection{Face-synthesis DeepFake}
\begin{figure}[!t]
\centering
\includegraphics[width=0.9\linewidth]{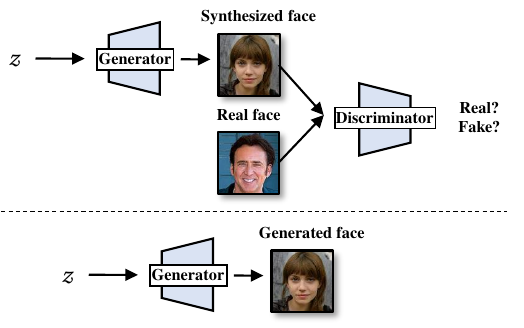}
\caption{\small The training phase (top) and testing phase (bottom) of the face-synthesis DeepFake model. $z$ denotes random noises.}
\vspace{-0.5cm}
\label{fig:face-synthesis}
\end{figure}

\vspace{-0.1cm}
\label{sec:fsyn}
With the popularization of AI techniques, DeepFake has become a more generalized term for AI-based face forgery techniques. Besides the face-swap technique, {another direction} is based on GAN models, which can create new faces that do not exist in reality by only giving random noises, \eg, \cite{goodfellow2014generative,karras2018progressive,karras2019style,peng2019cgr,karras2020analyzing}.
In general, a GAN model contains a generator and a discriminator. During training, the generator aims to create a new face from random noise, and the discriminator is used to distinguish the authenticity of the generated faces. The generator and discriminator combat each other until a balance is reached. In the testing phase, only the generator is used to map the random noise to a new face with high realism. The architecture, training, and generation process of face-synthesis DeepFake are shown in Fig.~\ref{fig:face-synthesis}.


\vspace{-0.3cm}
\subsection{DeepFake Detection}
\vspace{-0.1cm}
DeepFake detection attempts to determine the authenticity of given faces based on different feature signals, such as physiological signal \cite{agarwal2019protecting}, synthesis artifacts \cite{li2020face}, domain transformation \cite{li2020identification,qian2020thinking}, and data-driven training with various strategies, such as new designed architectures \cite{afchar2018mesonet,luo2021generalizing,zhao2021multi}, augmentations \cite{wang2021representative} and preprocessing \cite{nataraj2019detecting}. Specifically, the physiological signals used for DeepFake detection include speaking-action pattern \cite{agarwal2019protecting}, heartbeat rhythm \cite{hernandez2020DeepFakeson} and the combinations of multiple physiological signals \cite{ciftci2020fakecatcher}. For synthesis artifacts, Face X-ray \cite{li2020face} and Self-consistency \cite{zhao2021learning} improve the simulation process, leading to a large performance boost. Gu \etal \cite{gu2022exploiting} propose a progressive enhancement learning framework to utilize both the RGB and fine-grained frequency clues for face forgery detection. For data-driven training, Luo \etal \cite{luo2021generalizing} utilize high-frequency noise to improve cross-database scenarios. Zhao \etal \cite{zhao2021multi} utilize multiple spatial attention heads and textural feature enhancement blocks to boost performance. RFM \cite{wang2021representative} proposes an attention-based data augmentation framework to guide the detector to refine and enlarge its attention to boost detection performance. 
{Asnani \etal \cite{asnani2021reverse} propose fingerprint estimation and model parsing from generated images to boost detection performance. ITSNet \cite{wu2023interactive} explores the discriminant inconsistency representation by considering both RGB and frequency domain. Guo \etal \cite{guo2023rethinking} combine a gradient operator with CNNs to highlight the manipulation
traces left by face forgeries and further improve detection performance. Passive DeepFake defense attracts a lot of attention due to its good performance on multiple datasets. However, the passive DeepFake defense can only be used after DeepFake generation, which cannot obstruct the process of DeepFake generation.}

\vspace{-0.5cm}
\subsection{Proactive DeepFake Defense}
\vspace{-0.1cm}
In general, proactive methods tend to add marks in face images in advance, so as to disrupt the generation of DeepFakes. As of today, there are a few proactive DeepFake defense methods, \eg, \cite{segalis2020ogan,yeh2020disrupting,ruiz2020disrupting,wang2021faketagger,yang2021faceguard,yu2021artificial,asnani2022proactive}.

FaceGuard\cite{yang2021faceguard} designs a pattern of traces that are added to face images uploaded on social networks, and the traces can be eliminated in the DeepFake generation process. Then they detect whether the face images contain these traces. The face is regarded as real if it contains traces. Despite this method provides a concept for proactive DeepFake defense, it is not applicable, because real clean\footnote{{Clean means real face images not being processed by any method.}} faces and DeepFake faces can not be distinguished since both can contain no traces. 
In contrast, FakeTagger\cite{wang2021faketagger} encodes traces into face images and retains the traces after the DeepFake generation process. In other words, the generated faces would contain the traces. Note that the defined traces has a specific meaning. Then the DeepFake face can be detected if the traces can be correctly interpreted.
{Proactive-IMD\cite{asnani2022proactive} focuses on detecting images manipulated by GAN models. They do not interfere with the training of generative models. They propose to learn a set of templates adding onto real images. Their detection module could better discriminate the real images with templates and their manipulated counterparts.}
{Disrupting-DeepFakes\cite{ruiz2020disrupting} disrupt DeepFakes by adapting adversarial attack methods to image translation networks. They do not interfere with the training process of generative models. By adding perturbations on input images, the image translation networks could only generate images with bad visual quality.
}
{Similarly, the work of \cite{yeh2020disrupting} also applies an adversarial attack on images to cause the image translation model to fail in converting an image to the model’s designed outcome. They design two strategies to make the generative model either output a similar or unmodified version of the input image or output a broken and disfigured image.}
{OGAN\cite{segalis2020ogan} improves \cite{yeh2020disrupting} by proposing the Oscillating GAN(OGAN) attack method.}
These aforementioned works are commonly designed for the testing (generation) phase of the DeepFake model. They all regard the DeepFake model as one type of data perturbation process, which thereby does not affect the DeepFake model. {Moreover, these methods cannot simply use clean images as input for a generation. Instead, they require all input faces for testing to be processed.}
Fingerprint \cite{yu2021artificial} considers manipulating the training process of models. They train the model with images all being added on specific traces. In this way, the generated images would also be with traces. However, this method mainly focuses on GAN models, \ie, face-synthesis DeepFake. 
Since there is a large gap between face-swap DeepFake and face-synthesis DeepFake,  it does not apply well on face-swap DeepFakes. Moreover, Fingerprint is designed for model inventors instead of protecting users, as it can not distinguish real faces from traces and fake faces.

 These above works either focus on the testing phase, or the training stage of the face-synthesis DeepFake model. Still, they lack exploration on manipulating the training of face-swap DeepFake models. \blue{Fig. \ref{fig:face-swap} and Fig. \ref{fig:face-synthesis} exhibit the difference between face-synthesis and faces-swap DeepFakes tasks. There are significant differences in their application scenarios, as well as in their input forms during the training and testing phases.} In this paper, we focus on defending against the face-swap DeepFake by only implanting specific traces into training images.

\begin{figure*}[!ht]
\centering
\includegraphics[width=0.95\linewidth]{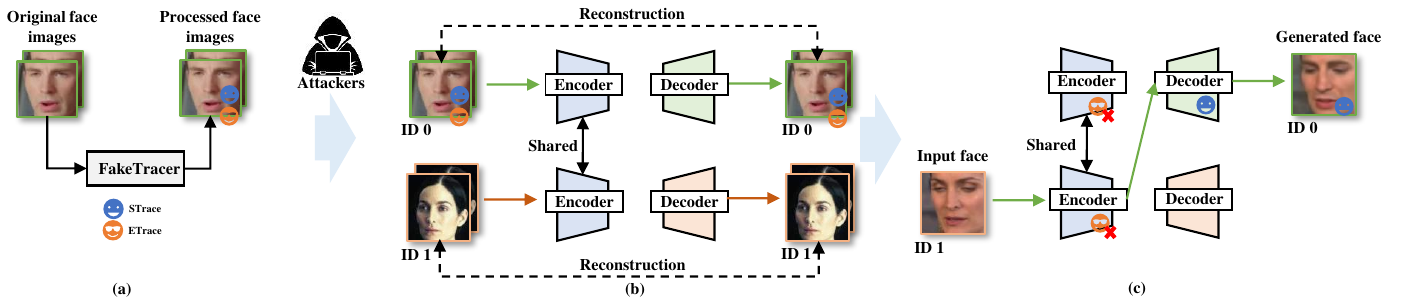}
\caption{\small Overview of our method. (a) The face images are implanted traces by our method before uploading them into the Internet. (b) The attackers collect the face images of a target identity (\eg, ID 0) to train the face-swap DeepFake model. Note that the training faces of another identity remain untouched. This setting is consistent with reality, as the users are expected to protect themselves by only modifying their face images. After training, the DeepFake model is learned to retain STrace but remove ETrace. (c) In the generation phase, given an input face of a source identity (\eg, ID 1), the generated face can contain STrace.}
\vspace{-0.6cm}
\label{fig:method}
\end{figure*}

\vspace{-0.1cm}
\section{Problem Settings}
\label{sec:problem}
The attackers can collect target face images from the Internet and use these images to train a face-swap DeepFake model. Then they can input the faces of the source identity to generate the faces of the target identity. To obstruct the generation of DeepFake, this paper proposes a new method to disrupt the DeepFake model by only manipulating the faces used for training. The overview of our method is shown in Fig.~\ref{fig:method}.
\vspace{-0.3cm}
\subsection{Problem Formulations}
\vspace{-0.1cm}
Denote $\mc{G}_{\theta}: \mc{X} \rightarrow \mc{X}$ as the mapping function of DeepFake model $\mc{G}$ with parameters $\theta$, where $\mc{X} = [0, 255]^{h \times w \times 3}$ is the image space. Given an input face image $x \in \mc{X}$, the generated face can be denoted as $x' = \mc{G}_{\theta}(x)$\footnote{We omit the notation of parameters for simplicity hereafter.}. Let $\mc{T}_s=\mc{H}_{s}(x)$ and $\mc{T}_e=\mc{H}_{e}(x)$ denote Sustainable Trace (STrace) and Erasable Trace (ETrace), and corresponding trace generators respectively. Our goal is to manipulate the training of the DeepFake model by adding these two traces on the training faces without any other intervention to the training process. Note that $\mc{T}_s$ should be embedded into the generation process while $\mc{T}_e$ should be eliminated by the generation process. By adding these two traces, the DeepFake model is learned only to insert STrace on generated faces. Denote $\mc{F}_s$ and $\mc{F}_e$ as the trace interpreter of traces $\mc{T}_s$ and $\mc{T}_e$ respectively. Exposing DeepFakes can be formulated as $\mc{F}_s(x') = \mc{T}_s$ and  $\mc{F}_e(x') \neq \mc{T}_e$.

\vspace{-0.3cm}
\subsection{Defense Workflow}
\label{sec:workflow}
To expose DeepFakes, the trace generators $\mc{H}_{s},\mc{H}_{e}$ and trace interpreters $\mc{F}_s,\mc{F}_e$ are provided to users as priors.  Then the users can create STraces and ETraces, and implant them to the face images uploaded to the Internet (See Fig.~\ref{fig:method}(a)). Once the attackers train the DeepFake model using these face images, the DeepFake model will be manipulated to generate faces containing STrace but no ETrace. {\bf Note that before training, only the face images of the target identity (\ie, the user is willing to be protected) are added these traces, and the training faces of source identity remain untouched. This setting is consistent with reality, as the users are expected to protect themselves by only modifying their face images} (See Fig.~\ref{fig:method}(b,c)). Then we can identify the authenticity of faces using the following criterion: we regard the faces as fake if STrace is detected, and the face as real if no traces are detected, or both STrace and ETrace are detected(Table \ref{tab:judge}).

\vspace{-0.3cm}
\begin{table}[!ht]
    \vspace{-0.2cm}
    \centering
    \small
    \caption{\small The criterion of the determination of authenticity.}
    \label{tab:judge}
    \vspace{-0.2cm}
    \begin{tabular}{cccccc}
       \toprule
       STrace & ETrace & Generated? & Remark \\
       \midrule
        \XSolidBrush & \XSolidBrush & No & Clean images \\
        \Checkmark & \Checkmark & No & FakeTracer processed images \\
        \Checkmark & \XSolidBrush & Yes & FakeTracer identified images \\
       \bottomrule
    \end{tabular}
    \vspace{-0.4cm}
\end{table}


\vspace{-0.5cm}
\subsection{Challenges}
\label{sec:challenges}
\vspace{-0.1cm}
The scenario in our method is more challenging compared to previous methods \cite{yang2021faceguard, wang2021faketagger, asnani2022proactive, yeh2020disrupting, ruiz2020disrupting, segalis2020ogan}, as we manipulate the training process instead of the testing process, in order to affect the generation of DeepFake model from the root, and the manipulation in this phase is restricted to the training face images. Some other methods focus on defending against face-synthesis DeepFake by manipulating the training process \cite{yu2021artificial}. However, these methods can hardly be employed to defend against face-swap DeepFake, as face-synthesis DeepFake focuses on face generation, which can create new faces based on random noises without considering the encoding process in identity switching. Thus their generator can be easily affected by the manipulated training face images. We take Fingerprint \cite{yu2021artificial} as an example for illustration. By inserting a specific pattern of traces into training faces, the generator can be guided to learn the mapping function from the random noises to the faces with traces added on.
However, the case is different in defending against face-swap DeepFake, as it takes face images as input instead of random noises and has an encoding process in training, which likely disrupts the traces implanted on training faces. Moreover, the training process of face-swap DeepFake is more complicated, as it utilizes an unsupervised training scheme involving two identities switching. Therefore, defending against face-swap DeepFake is more challenging.

\vspace{-0.3cm}
\section{Preliminary Validation}
\vspace{-0.1cm}
\label{sec:valid}

\begin{figure}[t]
\vspace{-0.4cm}
\centering
\includegraphics[width=0.8\linewidth]{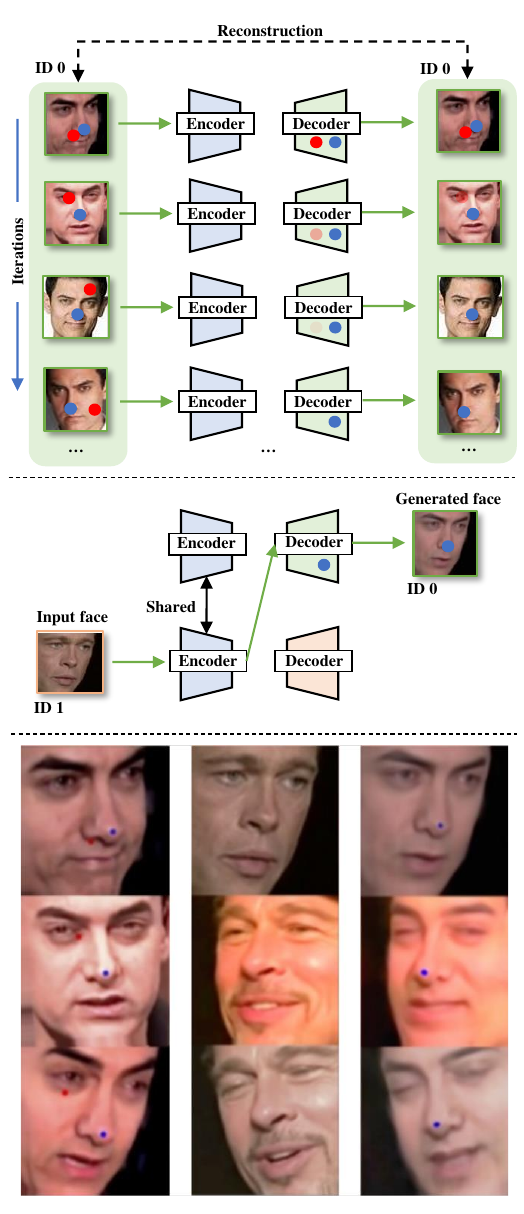}
\vspace{-0.5cm}
\caption{\small The visual illustration of our preliminary validation. (top) The illustration of the training process using faces of ID 0 with blue and red circles. (middle) In the generation phase, the clean face of ID 1 is selected as the ingredient to generate a face of ID 0. (bottom) The left column shows the training face images of ID 0 with blue and red circles. The middle column shows the input faces of ID 1 used for face swapping DeepFake. The right column corresponds to the generated faces of ID 0. }
\vspace{-0.6cm}
\label{fig:valid}
\end{figure}

To demonstrate the feasibility of our idea, we conduct a preliminary validation on DeepFake model using two identities of faces for training. Concretely, we design a blue circle, which is on the tip of nose on faces, determined by the facial landmarks. The blue circle is stable in the same relative position of all faces. Thus it is likely regarded as a part of face attributes that are expected to be retained by the encoder and learned by the decoder. The blue circle is analog to the STraces that can be retained in DeepFake generation. Analog to the ETrace, we design a red circle, which is added on a random face position. Due to the randomness, the red circle is irrelevant to the face content, thereby likely viewed as a part of the background that the encoder will ignore. 
Denote the source faces as ID 0, and the target faces as ID 1. 
We only manipulate the faces of ID 0 but keep the faces of ID 1 unchanged. 
As shown in Fig.~\ref{fig:valid}(top), we add these two circles on the training face images of ID 0. During the training, the blue circle is captured by the  decoder and can generate faces with this mark. In contrast, the red circle is gradually viewed as background and wiped off by the  encoder. After training, a clean face of ID 1 is selected as the ingredient fed into the  encoder and corresponding  decoder  to generate a face of ID 0. Since the decoder learns the pattern of the blue circle, it can generate faces with this mark, see Fig.~\ref{fig:valid}(middle). Several visual examples are shown in Fig.~\ref{fig:valid}(bottom). The left column is the training face images of ID 0 with blue circles on the tip of noises and red circles on random face positions. The middle column shows the input faces of ID 1 for face swapping. The right column corresponds to the generated faces of ID 0. We can observe that the DeepFake model can learn to perfectly retain the blue circle on the tip of the nose while removing the red circle on the faces, which can preliminarily prove that the DeepFake model can be affected by only adding these two marks on training faces.  
However, the blue and red circles are visible, hindering the practical use of defense. Thus in this paper, we intend to design less noticeable traces to substitute the blue and red circles.


\vspace{-0.3cm}

\section{Sustainable Traces}
\vspace{-0.1cm}
\label{sec:strace}

\begin{figure*}[!t]
\centering
\includegraphics[width=0.85\linewidth]{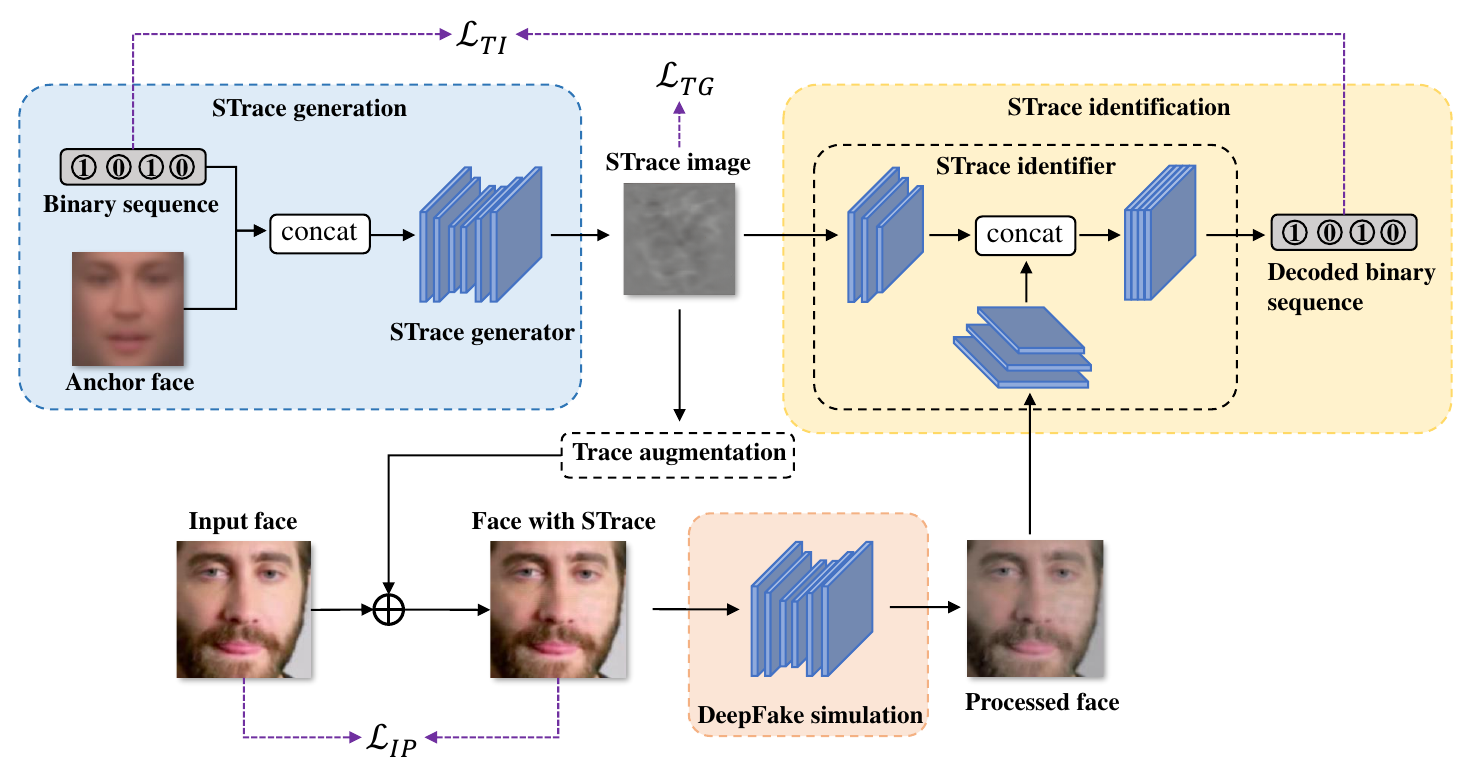}
\vspace{-0.4cm}
\caption{\small Overview of STrace generation, STrace identification, DeepFake simulation, and training objectives. See text for more details.}
\vspace{-0.5cm}
\label{fig:STrace}
\end{figure*}

The Sustainable Traces (STrace) implanted on training faces should meet the following three criteria: 1) have minimal effect on the visual quality of face images, 2) can be identifiable, and 3) can be embedded into DeepFake model. To achieve this goal, we develop a STrace generator to create traces, a STrace identifier to decode the traces, and a DeepFake simulator to imitate the process of face swapping, and design objectives to train these architectures jointly. 
The overview of creating STrace is shown in Fig.~\ref{fig:STrace}.
\vspace{-0.5cm}
\subsection{STrace Generation}
\vspace{-0.1cm}
The STrace generator is an encoder-decoder network where the input is an anchor face image and a binary input sequence, and the output is the image of STrace. The anchor face image is created by averaging all the faces in the training set, which is used to reduce the variance of generated traces.  The binary sequence corresponds to the generated traces, which is arbitrarily predefined by users. This network {consists of $10$ convolutional layers and $4$ upsampling layers. A ReLU function except the last one follows each convolutional layer.} 
Let the STrace generator as $\mc{H}_s$, the anchor face image as $\hat{x}$, and the input sequence as $v$. The generated STrace can be defined as $\mc{T}_s = \mc{H}_s(\hat{x}, v)$.

\vspace{-0.4cm}
\subsection{STrace Identification} 
\vspace{-0.1cm}
Given a face, we should identify the STrace to determine its authenticity. The STrace is successfully identified if the predefined binary sequence is recovered from the given face image. By doing so, we develop a STrace identifier to interpret the STrace from faces. Specifically, the STrace identifier is a two-branch network. One branch is the prior encoder, which corresponds to the input of generated STrace, and the other is the image encoder, which corresponds to the input of a given face. As shown in Fig.~\ref{fig:STrace}, one branch encodes the STrace into a vector, and the other encodes the given face into another vector. These two vectors are concatenated and forwarded to a sub-network to recover the binary sequence. Concretely, the STrace identifier {contains $7$ convolutional layers and $1$ fully-connected layer in the prior encoder, $7$ convolutional layers and $1$ fully-connected layer in image encoder, and $3$ fully-connected layers in sub-network. A ReLU function follows each convolutional layer.} It is noteworthy that the STrace generator is given to users in advance. Thus the users can know the pattern of STrace before identification, which is thereby used as an auxiliary to improve identification accuracy.  
Denote the STrace identifier as $\mc{F}_s$. Given a generated face $x'$, the recovered binary sequence can be defined as $v^{'} = \mc{F}_s(x', \mc{T}_s)$.
\vspace{-0.7cm}
\subsection{DeepFake Simulation} 
\vspace{-0.1cm}
We aim to manipulate the DeepFake model to generate faces with predefined binary sequences. The optimal solution for this goal is to train the STrace generator and identifier with the help of the face-swap DeepFake models. However, it is usually impractical to access the details of the DeepFake model in the wild. Therefore, we develop a substitute model to imitate the generation process of face-swap DeepFake. Considering the generation of face-swap DeepFake involves two key aspects, the encoding-decoding process and the identity switch. Compared to the identity switch, the encoding-decoding process has a significant impact on the STrace, as the STrace should survive the encoding process and be learned by the decoding process. To simulate this process, we develop a DeepFake simulator, which is an auto-encoder network similar to the face-swap DeepFake model, but is used only for face reconstruction. Despite not involving face swapping, it can well simulate the encoding-decoding process in the face-swap DeepFake model. 
The simulator is trained using self-collected faces in advance and then is deployed to reconstruct the faces added on STrace. With the help of this simulator, the STrace identifier can be promoted to learn the informative clues in STrace. DeepFake Simulator's encoder consists of $6$ convolutional layers, each followed by a LeakyReLU function; Its decoder consists of $5$ upsampling layers, $1$ convolutional layer and $1$ Sigmoid layer. There are $2$ linear layers and $1$ upsampling layer between the \blue{DeepFake Simulator's encoder  and decoder}.
\vspace{-0.4cm}
\subsection{Objectives and Training} 
\vspace{-0.1cm}
We jointly train the STrace generator and STrace identifier to achieve corresponding functions. Specifically, we develop three loss terms, which are trace generation loss $\mc{L}_{TG}$, image perceptual loss $\mc{L}_{IP}$, and trace identification loss $\mc{L}_{TI}$, respectively.
The overall objective function can be written as 
\begin{equation}
    \mc{L} = \lambda_1 \mc{L}_{TG} + \lambda_2 \mc{L}_{IP} + \lambda_3 \mc{L}_{TI}.
\end{equation}
The trace generation loss aims to minimize the magnitude of generated noise map as $\mc{L}_{TG} = || \mc{H}_s(\hat{x}, v) ||_{2}$. The image perceptual loss is to maintain the quality of face images after adding traces. We use LPIPS metric \cite{zhang2018unreasonable} in the experiment. The trace interpretation loss is used to recover the input sequence back from the generated faces, which is defined as cross-entropy loss $\mc{L}_{TI} = - \sum_{i} v_i \log (v'_i) + (1 - v_i) \log (1 - v'_i)$, where $i$ denotes the position of the sequence. $\lambda_1,\lambda_2,\lambda_3$ are weight factors to balance each loss term.

\vspace{-0.3cm}
\subsection{STrace Augmentation}
\vspace{-0.1cm}
Note that our method is applied in the setting where only the training data can be manipulated in the training process of the DeepFake model. Since the encoding process can naturally disrupt the implanted traces on training faces, we propose augmentation strategies to improve the resistance of STrace against the encoding process. Specifically, we randomly set $15\%-30\%$ pixels of trace map $\mc{T}_s$ as zero and multiply a scaling factor ranging from $0.8$ to $1.0$. After the trace is added to the input face, we apply other augmentations such as warping, blurring, noise adding, and JPEG compression to increase the diversity further.

\section{Erasable Traces}
\label{sec:etrace}
\begin{figure}[!ht]
\centering
\includegraphics[width=0.8\linewidth]{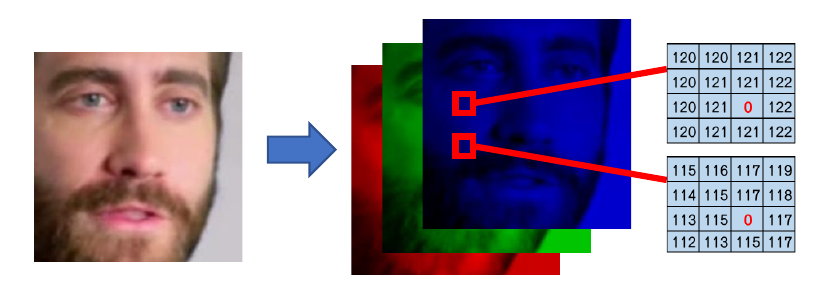}
\vspace{-0.4cm}
\caption{\small Illustration of adding Erasable Trace.}
\label{fig:09_traceB}
\end{figure}

Erasable Trace (ETrace) is proposed to help determine the authenticity of real faces not processed by our method. Specifically, the design of ETraces should also meet three criteria: 1) have minimal effect on the visual quality of face images, 2) can be identifiable, and 3) can be discarded in the DeepFake generation process. To achieve this goal, we create traces independent of the face content, based on the validation in Section \ref{sec:valid} that the \blue{DeepFake decoder} can not learn a pattern irrelevant to the face attributes. As shown in Fig. \ref{fig:09_traceB}, we select a set of positions on the image and set a predefined value to the blue channel of these positions. These positions and predefined values are prior information known to users. Thus the trace can be inserted into any position with any value. This process is imperceptible to human observers, but it can be easily identified if this prior information is provided.

\vspace{-0.5cm}
\section{Experiment}
\label{sec:exp}
\vspace{-0.1cm}
\subsection{Experimental Settings}
\vspace{-0.1cm}
\noindent{\bf Dataset.}
\label{dataset}
We use DeepFake dataset Celeb-DF \cite{li2020celeb} for experiments. For embedding sustainable traces,  we randomly select 5000 real faces covering all the identities for training and another 5000 real faces for testing. All the faces have the size of $256 \times 256$. 


\smallskip
\noindent{\bf DeepFake Model.}
We employ the DeepFake model proposed in Celeb-DF \cite{li2020celeb} as a demonstration. Specifically, we follow the same training protocol that one encoder and $n$ decoders for different identities are trained alternatively. For simplicity, we use $n = 2$ identities. Note that in each training pair, only the target identity's faces are added sustainable and erasable traces, while the faces of the source identity are clean.

\smallskip
\noindent{\bf Metrics.} \blue{We use five metrics for evaluation: B-Acc, SSIM, PSNR, FID, and D-Acc.}
(1) B-Acc is the bit accuracy between the decoded binary sequence and the original input sequence, which indicates how well STrace is retained after DeepFake generation. (2) \blue{SSIM, PSNR, and FID are calculated between clean images and images processed with our method. They are used to evaluate the quality of faces with traces on them.} (3) D-Acc stands for detection accuracy, which indicates whether a face is correctly classified as real or generated. The face is considered fake if STrace is detected and no ETrace is detected (As shown in Table \ref{tab:judge}). The face contains STrace if B-Acc is larger than a threshold. \footnote{\blue{We refer to B-Acc on generated images, D-Acc on generated images, and D-Acc on real images as B-G, D-G, and D-R, respectively.}}

\smallskip
\noindent{\bf Compared Methods.} To demonstrate the efficacy of our method, we employ two passive detection methods, CNND \cite{wang2020cnn} and RFM\cite{wang2021representative}, and proactive method Fingerprint \cite{yu2021artificial} for comparison. We use the default settings provided by the authors.

\smallskip
\noindent{\bf Implementation Details.}
Our method is implemented using PyTorch 1.7.1 on Ubuntu 18.04 with an Nvidia 3090 GPU. In the experiments, we set the length of the binary sequence as $8$ bits. The batch size is set as $4$, initial learning rate, $\lambda_{1}, \lambda_{2}, \lambda_{3}$    are set as $2.5\times10^{-5}, 2.0, 2.0, 6.0$. For optimizer we utilize Adam optimizer. We test the decoded message accuracy every 5000 iterations. If the accuracy is higher than a threshold(set as $0.89$), we multiply $\lambda_{1}, \lambda_{2}$ by $1.25$; If the accuracy is lower than the threshold, we multiply $\lambda_{3}$ by $1.25$. The B-Acc threshold for deciding whether the face is with STrace in D-Acc is $75\%$.

\vspace{-0.5cm}
\subsection{Results}
\label{sec:results}
\vspace{-0.1cm}

\noindent{\bf B-Acc on generated Faces.}
We train the DeepFake model for $6$ different pairs of identities and each identity uses around 4000 faces for training.
Bit Accuracy for the generated images (B-Acc on G) of different pairs of identities is shown in Table \ref{tab:00_bacc}. The $\rightarrow$ indicates the source identity to the target identity. We can observe that our method can interpret the binary sequence from the generated faces precisely (over $90\%$), which indicates that the sustainable traces can resist the generation process in the DeepFake model.

\begin{table}[!ht]
    \vspace{-0.5cm}
    \centering
    \small
    \caption{\small B-Acc on generated faces.}
    \label{tab:00_bacc}
    \vspace{-0.2cm}
    \begin{tabular}{c|c|c|c|c|c|c}
    \toprule
    \multirow{3}{*}{ID pairs}
    & 59 & 57 & 56 & 54 & 45 & 42 \\
    & $\downarrow$  & $\downarrow$ & $\downarrow$  &  $\downarrow$ & $\downarrow$ & $\downarrow$ \\
    & 2  &  4 & 5  &  7 & 16 & 19 \\
    \midrule
    B-Acc on G & 0.93 & 0.93 & 0.92 & 0.97 & 0.94 & 0.92 \\
    \bottomrule
    \end{tabular}
\end{table}



\noindent{\bf D-Acc and SSIM.}
To obtain D-Acc, we need to tell whether the face contains STrace and ETrace. The detection of ETrace is accurate, which can achieve $100\%$ accuracy on both real and generated images. Thus we only need to decide whether the face contains STrace. 
Table \ref{tab:01_detect_acc} shows the performance of our method on averaging six identities with comparison to the state-of-the-art passive detection methods CNND \cite{wang2020cnn}, RFM\cite{wang2021representative}, and proactive method Fingerprint \cite{yu2021artificial} \blue{and Proactive-IMD\cite{asnani2022proactive}. From results, we can observe that Fingerprint and Proactive-IMD are not effective to defend against face-swap DeepFake compared to us.} Moreover, our method can achieve competitive performance with RFM and outperforms CNND by a large margin. Additionally, the $0.89$ SSIM score indicates the traces have 
a small visible effect on the image quality.

\begin{table}[!ht]
    \small
    \centering
    \vspace{-0.5cm}
    \caption{\small Performance of different methods.}
    \label{tab:01_detect_acc}
    \vspace{-0.2cm}
    \begin{tabular}{c|c|c|c|c|c}
    \toprule
    Methods	& \blue{D-G}  & \blue{D-R} & SSIM & \blue{PSNR} & \blue{FID} \\
    \midrule
    CNND\cite{wang2020cnn} & 0.03 & 0.99 & / & / & / \\
    \midrule
    RFM\cite{wang2021representative} & 0.99 & 0.82 & / & / & / \\
    \midrule
    Fingerprint \cite{yu2021artificial} & 0.0 & 1.0 & 0.99 & \blue{47.54} & \blue{3.36} \\
    \midrule
    \blue{Proactive-IMD \cite{asnani2022proactive}} & \blue{0.0} & \blue{1.0} & \blue{0.98} &\blue{42.95} & \blue{38.97} \\
    \midrule
    \bf FakeTracer & 0.95 & 0.84 & 0.89 & \blue{36.22} & \blue{21.91}\\
    \bottomrule
    \end{tabular}
\end{table}

\begin{figure*}[t]
        \centering
        \vspace{-0.3cm}
        \includegraphics[width=\linewidth]{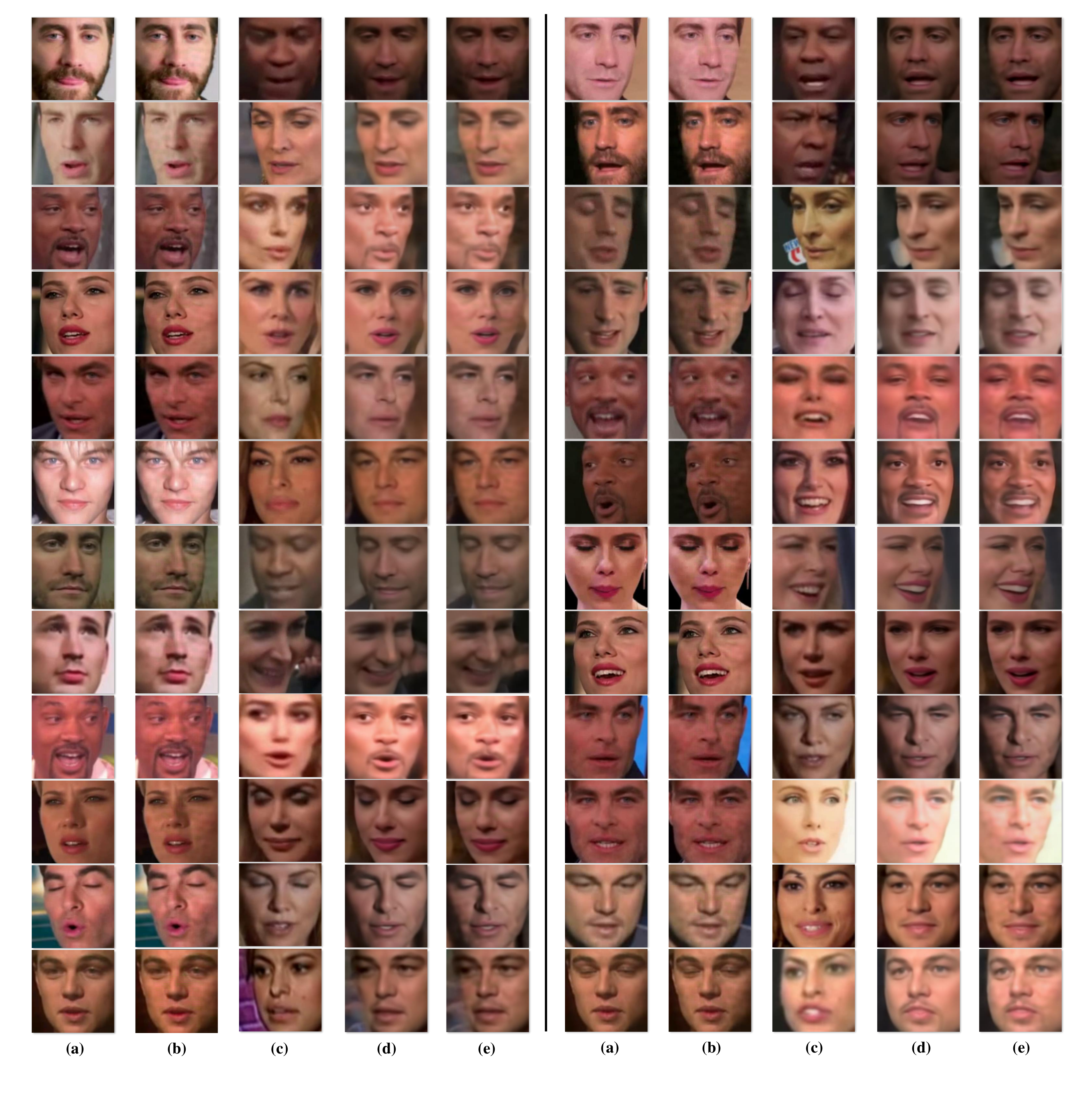}
        \vspace{-1.2cm}
        \caption{\small Visual examples. (a) Clean real faces. (b) Faces with STrace and ETrace. (c) DeepFake Model inputs (Testing Phase). (d) Generated faces trained with clean faces (Testing Phase). (e) Generated faces trained with processed faces (Testing Phase). }
        \label{fig:05_visual}
        \vspace{-0.5cm}
\end{figure*}

\noindent{\bf Visual Examples.} 
Fig.\ref{fig:05_visual} illustrates several visual examples created by our method. We could observe that traces added by our method have less effect on image quality (Fig.\ref{fig:05_visual}(a, b)), and generated faces with the traces have a similar quality to the ones generated by DeepFake in regular training (Fig.\ref{fig:05_visual}(e, d)).

\smallskip
\noindent{\bf User Study for Visual Perceptibility.}
{This part describes a user study for the visual perceptibility of faces processed by our method. Specifically, we design an online questionnaire to study whether the traces can influence identity recognition by human observers. For each question, we first demonstrate $5$ clean face images from the same identity and ask the volunteers to score how much they feel like these images belong to the same identity. The scores are ranging from $0$ to $10$. We then represent these face images with traces added on and ask the volunteers to repeat the assessment process. The smaller difference between these two scores represents a smaller influence caused by our method on human eyes. Concretely, we choose $10$ identities in each questionnaire and ask $20$ volunteers to score. The final score of the mean absolute value difference is $0.29$, which is quite small compared to the score ranging $[0, 10]$. This user study demonstrates that our method has little influence on visual perceptibility.}

\vspace{-0.4cm}
\subsection{Ablation Study}
\label{ablation}
\vspace{-0.1cm}
\noindent{\bf Effect of Different Components and Strategies.}
We explore the effects of different components or strategies for embedding STrace in Table \ref{tab:03_ablation}. Specifically, we study two strategies that are used for trace enhancement, which are random zero setting (RZS) and random scaling (RS). We could observe that the B-Acc notably drops while the SSIM score is improved, which is because the STrace generator learns to generate less intensive sustainable trace and those traces are missing on the generated faces. 
Moreover, we study the effect of the prior encoder and DeepFake Simulator (DFS) respectively.
From Table \ref{tab:03_ablation} we could observe that removing the prior encoder reduces B-Acc on generated images significantly, which demonstrates that the prior encoder is an effective component for decoding sequence. We can also observe that DeepFake Simulator is critical to the performance compared to using random zero setting and random scaling, which largely drops the B-Acc once it is removed.

\begin{table}[t]
    \vspace{-0.5cm}
    \centering
    \small
    \caption{\small Ablation Study. G is for generated face images, and R is for real face images, RZS is for Random Zero Setting, RS is for Random Scaling, DFS is for DeepFake Simulator.}
    \label{tab:03_ablation}
    \vspace{-0.2cm}
    \begin{tabular}{c|c|c|c|c}
    \toprule
    Methods	& B-Acc on G$\uparrow$ & SSIM$\uparrow$ & \blue{PSNR$\uparrow$} & \blue{FID$\downarrow$} \\
    \midrule
    All & 0.94 & 0.89 & \blue{36.22} & \blue{21.91}\\
    \midrule
    w/o RZS & 0.82 & 0.98 & \blue{40.14} & \blue{8.51}\\
    \midrule
    w/o RS & 0.69 & 0.94 & \blue{39.07} & \blue{10.23}\\
    \midrule
    w/o RZS + RS & 0.76 & 0.98 & \blue{42.16} & \blue{7.03}\\
    \midrule
    w/o prior encoder & 0.75 & 0.98 & \blue{38.92} & \blue{7.68}\\
    \midrule
    w/o DFS & 0.58 & 0.98 & \blue{44.12} & \blue{24.67}\\
    \bottomrule
    \end{tabular}
    \vspace{-0.5cm}
\end{table}

\begin{figure}[!ht]
        \centering
        \vspace{-0.3cm}\includegraphics[width=0.85\linewidth]{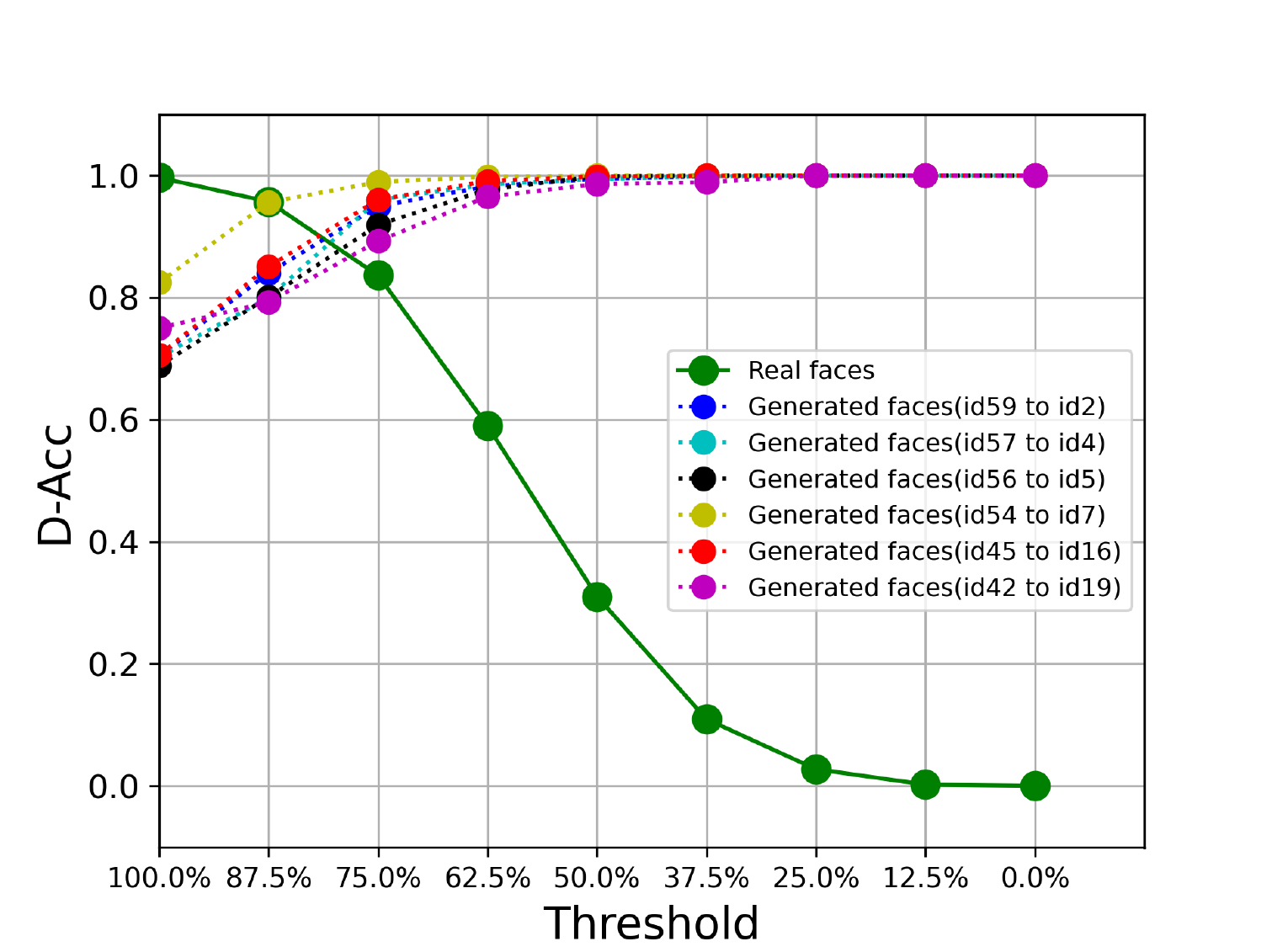}
        \caption{\small The relationship between D-Acc and B-Acc on real and generated faces.}
        \label{fig:08_thresh}
        \vspace{-0.5cm}
\end{figure}

\smallskip
\noindent{\bf The Effect of Different B-Acc threshold on D-Acc.}
{Fig.\ref{fig:08_thresh} illustrates D-Acc along with different values of B-Acc thresholds. Note that the threshold $100\%$ denotes the decoded binary sequence completely matching the input binary sequence. We can observe that the D-Acc on real images drops with threshold reducing, as the sequence with less bit matched can be treated as generated faces. On the contrary, D-Acc on generated faces increases with threshold reducing. Thus we select $75\%$ as the threshold for a better balance.}

\smallskip
\noindent{\bf The Effect of Different Fraction of Faces with Traces in Training.}
We are motivated to investigate the effect of using different fractions of faces with traces in training, as it is unlikely to implant traces on all training faces collected by attackers in real-world scenarios. Thus it is important to study whether the proactive defense can succeed when a small part of training images are processed. To this end, we conduct experiments using different fractions of faces with traces implanted. The results are shown in Table \ref{tab:05_fraction}, where $100\%$ represents that all the training faces are processed by our method and the same to the fraction of $75\%$ and $50\%$. 
From Table \ref{tab:05_fraction}, we can observe that our method remains effective even if a portion of training images are processed. Specifically, our method can still perform well at a fraction of $75\%$. Despite the performance of using $50\%$ training faces with traces being degraded, STrace can still be detected in over half ($0.58$) generated faces.

\vspace{-0.2cm}
\begin{table}[!ht]
    \small
    \centering
    \vspace{-0.2cm}
    \caption{\small Performance of different fractions of training images that are marked. G is for generated face images, and R is for real face images.}
    \label{tab:05_fraction}
    \vspace{-0.2cm}
    \begin{tabular}{c|c|c|c}
    \toprule
    Fraction & B-Acc on G & D-Acc on G  & D-Acc on R  \\
    \midrule
    $100\%$ & 0.94 & 0.95 & 0.84 \\
    \midrule
    $75\%$ & 0.85 & 0.80 & 0.84 \\
    \midrule
    $50\%$ & 0.73 & 0.58 & 0.84 \\
    \bottomrule
    \end{tabular}
    \vspace{-0.5cm}
\end{table}

\smallskip
\noindent{\bf The Effect of Different DeepFake Model Architectures.} \blue{This part investigates the generalization of our method toward different DeepFake models.} Note the default DeepFake model used in our experiments has an input size of $256 \times 256$ and an output size of $256 \times 256$. To increase the diversity of the DeepFake model, we create new architectures using the following rules. For \blue{DeepFake encoders}, we create two more variants that have input sizes of $128 \times 128, 64 \times 64$. Specifically, we add corresponding downsampling convolutional layers, which have kernel size of $5$ and stride of $2$. Similarly, we create two more variants for \blue{DeepFake decoders} which have input sizes of $128 \times 128, 64 \times 64$ as well. Then we create a new DeepFake model by randomly combining one \blue{DeepFake encoder} and a \blue{DeepFake decoder}. To fit in the newly created DeepFake model, our method downsamples the face image resolution without changing the architecture and training details. The performance of our method on different DeepFake models is shown in Table \ref{tab:06_size}. Note that $n \rightarrow m$ denotes that the DeepFake model consists of a \blue{DeepFake encoder} that has an input size of $n\times n$ and a \blue{DeepFake decoder} that has an output size of $m\times m$. On the one hand, the results illustrate that the detection performance is highly relevant to the \blue{DeepFake decoder} architecture. The larger the output size of the \blue{DeepFake decoder} is, the better the performance our method can achieve. On the other hand, the variance in the input size of \blue{DeepFake encoder} has little influence on the performance. This is mainly due to that a high capacity of the decoder can learn more clues of implanted traces, which thereby can generate faces with STrace even though the \blue{DeepFake encoder} is small. \blue{We also evaluate our method on DeepFaceLab\cite{liu2023deepfacelab}. Table \ref{tab:06_size} shows that our method also performs favorably on this architecture, showing well generalization ability.}
\begin{table}[!ht]
    \small
    \centering
    \vspace{-0.45cm}
    \caption{\small Performance of using different DeepFake model architectures. $n \rightarrow m$ denotes that the DeepFake model consists of an encoder that has an input size of $n\times n$ and a decoder that has an output size of $m\times m$.}
    \label{tab:06_size}
    \vspace{-0.2cm}
    \begin{tabular}{c|c|c|c}
    \toprule
    \blue{Architecture} & \blue{B-G} & \blue{D-G}  & \blue{D-R}  \\
    \midrule
    $256 \rightarrow 256$ & 0.94 & 0.95 & 0.84 \\
    \midrule
    $128 \rightarrow 256$ & 0.94 & 0.94 & 0.84 \\
    \midrule
    $64 \rightarrow 256$ & 0.91 & 0.92 & 0.84 \\
    
    \midrule
    $128 \rightarrow 128$ & 0.81 & 0.70 & 0.83 \\
    \midrule
    $256 \rightarrow 128$ & 0.83 & 0.74 & 0.83 \\
    \midrule
    $64 \rightarrow 64$ & 0.73 & 0.55 & 0.86 \\
    \midrule
    $256 \rightarrow 64$ & 0.73 & 0.55 & 0.86 \\
    \midrule
    \midrule
    \blue{DeepFaceLab\cite{liu2023deepfacelab}} & \blue{0.89} & \blue{0.87} & \blue{0.84} \\
    
    \bottomrule
    \end{tabular}
\end{table}

\smallskip
\noindent{\blue{\textbf{Evaluation on other datasets.}}  \blue{We evaluate our method on another two DeepFake datasets, DFDC and FaceForensics++. We select 6 pairs of identities from each dataset and repeat the same evaluation method. The results are shown in Table \ref{tab:07_detect_acc_all}. It can be observed that the performance of our method is stable across different datasets.} 

\begin{table}[!ht]
    \small
    \centering
    \vspace{-0.5cm}
    \caption{\small \blue{Performance on different datasets.}}
    \label{tab:07_detect_acc_all}
    \vspace{-0.2cm}
    \begin{tabular}{c|c|c|c|c|c|c}
    \toprule
    \blue{
    Datasets}& \blue{B-G} & \blue{D-G} & \blue{D-R} & \blue{SSIM} & \blue{PSNR} & \blue{FID} \\
    \midrule
    \textbf{\blue{Celeb-DF}} & \blue{0.94} & \blue{0.95} & \blue{0.84} & \blue{0.89} & \blue{36.22} & \blue{21.91}\\
    \midrule
    \textbf{\blue{DFDC}} & \blue{0.92} & \blue{0.90} & \blue{0.81} & \blue{0.90} & \blue{36.23} & \blue{24.25}\\
    \midrule
    \textbf{\blue{FF++}} & \blue{0.92} & \blue{0.92} & \blue{0.82} & \blue{0.88} & \blue{36.23} & \blue{31.43}\\
    \bottomrule
    \end{tabular}
    \vspace{-0.7cm}
\end{table}

\vspace{-0.3cm}
\subsection{Robustness}
\vspace{-0.1cm}
The robustness of our method is essential for real-world applications, as the generated faces possibly go through a variety of post-processing operations when uploading them to the Internet. To demonstrate the robustness of our method, we apply different types of post-processing operations to the generated faces and the faces processed by our method for training DeepFake and then evaluate the defense performance. This setting is practical as both the generated faces and the faces processed by our method for training DeepFake can be subject to these post-processing operations on the Internet. Specifically, we use three kinds of post-processing operations, which are JPEG compression (quality factor as $25$), Gaussian blurring (sigma $3.0$, kernel size $5 \times 5$), and Gaussian noise (sigma $0.1$, mean value $0.0$) respectively.

Table \ref{tab:02_robust} shows the D-Acc of our method against these post-processing operations {on generated faces. The training protocol is the same as Table \ref{tab:00_bacc}.} 
{Table \ref{tab:02_robust_train} shows the D-Acc of our method with these post-processing operations on the face images processed by our method for training the DeepFake model. The results in these tables reveal that the performance of our method only slightly drops with these post-processing operations, which is mainly because the strategies we develop for trace enhancement in training improve the ability of trace interpreters, and the proposed DeepFake Simulator improves the generalization of learning.}

\begin{table}[!ht]
    \small
    \centering
    \vspace{-0.5cm}
    \caption{\small Robustness of generated faces against three post-processing operations. JPEG, GB and GN represent JPEG compression, Gaussian blurring, and Gaussian noise, respectively.}
    \vspace{-0.2cm}
    \label{tab:02_robust}
    \begin{tabular}{c|c|c}
    \toprule
    Perturbations & D-Acc on G & D-Acc on R \\
    \midrule
    None & 0.95 & 0.84 \\
    \midrule
    JPEG & 0.93 & 0.84 \\
    \midrule
    GB & 0.92 & 0.84 \\
    \midrule
    GN & 0.95 & 0.84 \\
    \midrule
    \end{tabular}
\end{table}

\begin{table}[!ht]
    \vspace{-0.2cm}
    \small
    \centering
    \vspace{-0.4cm}
    \caption{\small Robustness of face images processed by our method against three post-processing operations. JPEG, GB and GN represent JPEG compression, Gaussian blurring, and Gaussian noise, respectively.}
    \vspace{-0.2cm}
    \label{tab:02_robust_train}
    \begin{tabular}{c|c|c}
    \toprule
    Perturbations & D-Acc on G & D-Acc on R \\
    \midrule
    None & 0.95 & 0.84 \\
    \midrule
    JPEG & 0.93 & 0.84 \\
    \midrule
    GB & 0.94 & 0.84 \\
    \midrule
    GN & 0.92 & 0.84 \\
    \midrule
    \end{tabular}
    \vspace{-0.6cm}
\end{table}

\vspace{-0.2cm}

\section{{Discussion}}
\vspace{-0.1cm}
\smallskip
\noindent{\bf The Role of Anchor Face in STrace Generation.} The anchor face in STrace generation aims to provide extra guidance to the encoding process from binary sequence to the trace map. Since the anchor face is obtained on average of many faces, it likely inserts face-relevant information into the generation, leading to a stable training of STrace.

\vspace{-0.1cm}
\smallskip
\noindent{\bf The Connection between Generation Ability and Defense Performance.}
Our method highly associates with the generation quality of the DeepFake model, \ie, the generation ability of the DeepFake model. Our method can perform well on powerful DeepFake models, as they are competent in learning the pattern of STrace. The case differs from compromised models as they can not generate faces well, much less the STrace. Nevertheless, low-quality generated faces always have apparent artifacts which humans can easily distinguish. Thus it is not necessary to defend against these compromised models. Fig.~\ref{fig:032103_arti} shows examples of faces generated by a DeepFake model trained with frontal faces. When given an input face with an extreme angle, the DeepFake model cannot generate a high-quality face, subsequently affecting the detection of STrace. 

\begin{figure}[!ht]
        \centering
        \vspace{-0.4cm}
        \includegraphics[width=0.4\linewidth]{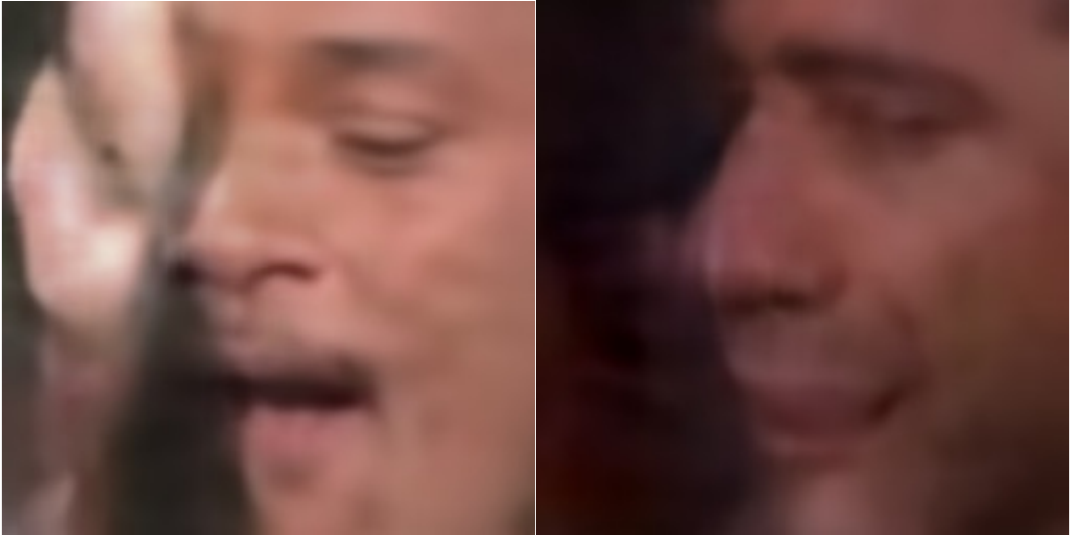}
        \caption{\small Examples of DeepFake generated faces by inputting faces with an extreme angle.}
        \label{fig:032103_arti}
        \vspace{-0.35cm}
\end{figure}

\noindent{\bf The Feasibility of Implanting Traces Locally.} It remains an open question of what kind of STrace can affect the DeepFake model best. In this part, we investigate the feasibility of implanting the traces into the local region. As shown in Fig. \ref{fig:032102_local}, we add traces only on the region around the nose, as this region is usually complex that can better hide traces. However, the trace becomes very intensive, largely downgrading the visual quality.

\begin{figure}[!ht]
        \centering
        \vspace{-0.4cm}
        \includegraphics[width=0.4\linewidth]{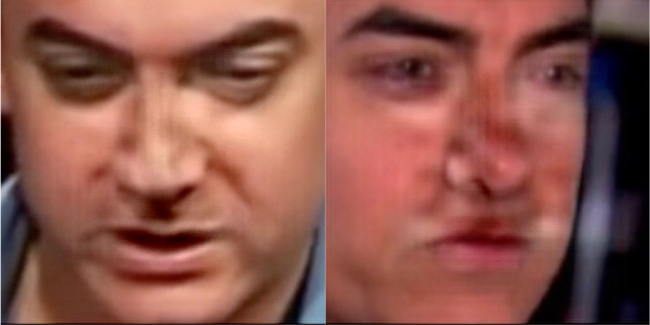}
        \caption{\small Examples of implanting STrace into the local region around the nose.}
        \label{fig:032102_local}
        \vspace{-0.5cm}
\end{figure}

\vspace{-0.2cm}
\section{Conclusion}
This paper describes a new method named FakeTracer to proactively defend against face-swap DeepFake faces. Specifically, we design two types of traces, sustainable trace and erasable trace, that can be added to the training faces to affect the training of DeepFake models. By training on these traced face images, the DeepFake model can learn to insert sustainable traces on generated faces while wiping off the erasable trace. Then we develop different interpreters to decode these traces from face images. Extensive experiments on the Celeb-DF dataset demonstrate the efficacy of our method.

\noindent{\bf Future Work.}
We would like to reduce the distortion of STrace further while increasing its load of information. Specifically, we aim to improve the architecture of the STrace generator and identifier with the newly proposed ViTs as the backbone network. Moreover, we intend to further improve the robustness of STrace by incorporating more effective enhancement strategies.

\smallskip
\noindent{\bf Acknowledgments.}
This material is based upon work supported by the China Postdoctoral Science Foundation under Grant No. 2021TQ0314 and 2021M703036, and the Fundamental Research Funds for the Central Universities.
\vspace{-0.5cm}
\bibliographystyle{IEEEtran}
\bibliography{ref}

\begin{thebibliography}{10}
\providecommand{\url}[1]{#1}
\csname url@samestyle\endcsname
\providecommand{\newblock}{\relax}
\providecommand{\bibinfo}[2]{#2}
\providecommand{\BIBentrySTDinterwordspacing}{\spaceskip=0pt\relax}
\providecommand{\BIBentryALTinterwordstretchfactor}{4}
\providecommand{\BIBentryALTinterwordspacing}{\spaceskip=\fontdimen2\font plus
\BIBentryALTinterwordstretchfactor\fontdimen3\font minus \fontdimen4\font\relax}
\providecommand{\BIBforeignlanguage}[2]{{%
\expandafter\ifx\csname l@#1\endcsname\relax
\typeout{** WARNING: IEEEtran.bst: No hyphenation pattern has been}%
\typeout{** loaded for the language `#1'. Using the pattern for}%
\typeout{** the default language instead.}%
\else
\language=\csname l@#1\endcsname
\fi
#2}}
\providecommand{\BIBdecl}{\relax}
\BIBdecl

\bibitem{goodfellow2014generative}
I.~Goodfellow, J.~Pouget-Abadie, M.~Mirza, B.~Xu, D.~Warde-Farley, S.~Ozair, A.~Courville, and Y.~Bengio, ``Generative adversarial nets,'' \emph{Advances in Neural Information Processing Systems (NeurIPS)}, vol.~27, 2014.

\bibitem{kingma2014auto}
D.~P. Kingma and M.~Welling, ``Auto-encoding variational bayes,'' \emph{stat}, vol. 1050, p.~1, 2014.

\bibitem{nguyen2019deep}
T.~T. Nguyen, Q.~V.~H. Nguyen, C.~M. Nguyen, D.~Nguyen, D.~T. Nguyen, and S.~Nahavandi, ``Deep learning for deepfakes creation and detection: A survey,'' \emph{arXiv preprint arXiv:1909.11573}, 2019.

\bibitem{liu2023deepfacelab}
K.~Liu, I.~Perov, D.~Gao, N.~Chervoniy, W.~Zhou, and W.~Zhang, ``Deepfacelab: integrated, flexible and extensible face-swapping framework,'' \emph{Pattern Recognition}, p. 109628, 2023.

\bibitem{gupta2023towards}
A.~Gupta, R.~Mukhopadhyay, S.~Balachandra, F.~F. Khan, V.~P. Namboodiri, and C.~Jawahar, ``Towards generating ultra-high resolution talking-face videos with lip synchronization,'' in \emph{IEEE Winter Conference on Applications of Computer Vision (WACV)}, 2023, pp. 5209--5218.

\bibitem{akhtar2023deepfakes}
Z.~Akhtar, ``Deepfakes generation and detection: A short survey,'' \emph{Journal of Imaging}, vol.~9, no.~1, p.~18, 2023.

\bibitem{pantserev2020malicious}
K.~A. Pantserev, ``The malicious use of ai-based deepfake technology as the new threat to psychological security and political stability,'' \emph{Cyber defence in the age of AI, smart societies and augmented humanity}, pp. 37--55, 2020.

\bibitem{zhao2021multi}
H.~Zhao, W.~Zhou, D.~Chen, T.~Wei, W.~Zhang, and N.~Yu, ``Multi-attentional deepfake detection,'' in \emph{IEEE Conference on Computer Vision and Pattern Recognition (CVPR)}, 2021, pp. 2185--2194.

\bibitem{li2020face}
L.~Li, J.~Bao, T.~Zhang, H.~Yang, D.~Chen, F.~Wen, and B.~Guo, ``Face x-ray for more general face forgery detection,'' in \emph{IEEE Conference on Computer Vision and Pattern Recognition (CVPR)}, 2020, pp. 5001--5010.

\bibitem{qian2020thinking}
Y.~Qian, G.~Yin, L.~Sheng, Z.~Chen, and J.~Shao, ``Thinking in frequency: Face forgery detection by mining frequency-aware clues,'' in \emph{European Conference on Computer Vision (ECCV)}.\hskip 1em plus 0.5em minus 0.4em\relax Springer, 2020, pp. 86--103.

\bibitem{luo2021generalizing}
Y.~Luo, Y.~Zhang, J.~Yan, and W.~Liu, ``Generalizing face forgery detection with high-frequency features,'' in \emph{IEEE Conference on Computer Vision and Pattern Recognition (CVPR)}, 2021, pp. 16\,317--16\,326.

\bibitem{li2021frequency}
J.~Li, H.~Xie, J.~Li, Z.~Wang, and Y.~Zhang, ``Frequency-aware discriminative feature learning supervised by single-center loss for face forgery detection,'' in \emph{IEEE Conference on Computer Vision and Pattern Recognition (CVPR)}, 2021, pp. 6458--6467.

\bibitem{qiao2018statistical}
T.~Qiao, R.~Shi, X.~Luo, M.~Xu, N.~Zheng, and Y.~Wu, ``Statistical model-based detector via texture weight map: Application in re-sampling authentication,'' \emph{IEEE Transactions on Multimedia (TMM)}, vol.~21, no.~5, pp. 1077--1092, 2018.

\bibitem{chen2020serial}
B.~Chen, W.~Tan, G.~Coatrieux, Y.~Zheng, and Y.-Q. Shi, ``A serial image copy-move forgery localization scheme with source/target distinguishment,'' \emph{IEEE Transactions on Multimedia (TMM)}, vol.~23, pp. 3506--3517, 2020.

\bibitem{asnani2021reverse}
V.~Asnani, X.~Yin, T.~Hassner, and X.~Liu, ``Reverse engineering of generative models: Inferring model hyperparameters from generated images,'' \emph{arXiv preprint arXiv:2106.07873}, 2021.

\bibitem{gu2022exploiting}
Q.~Gu, S.~Chen, T.~Yao, Y.~Chen, S.~Ding, and R.~Yi, ``Exploiting fine-grained face forgery clues via progressive enhancement learning,'' in \emph{AAAI Conference on Artificial Intelligence (AAAI)}, vol.~36, no.~1, 2022, pp. 735--743.

\bibitem{wu2023interactive}
J.~Wu, B.~Zhang, Z.~Li, G.~Pang, Z.~Teng, and J.~Fan, ``Interactive two-stream network across modalities for deepfake detection,'' \emph{IEEE Transactions on Circuits and Systems for Video Technology (TCSVT)}, 2023.

\bibitem{guo2023rethinking}
Z.~Guo, G.~Yang, D.~Zhang, and M.~Xia, ``Rethinking gradient operator for exposing ai-enabled face forgeries,'' \emph{Expert Systems with Applications}, vol. 215, p. 119361, 2023.

\bibitem{yang2021faceguard}
Y.~Yang, C.~Liang, H.~He, X.~Cao, and N.~Z. Gong, ``Faceguard: Proactive deepfake detection,'' \emph{arXiv preprint arXiv:2109.05673}, 2021.

\bibitem{wang2021faketagger}
R.~Wang, F.~Juefei-Xu, M.~Luo, Y.~Liu, and L.~Wang, ``Faketagger: Robust safeguards against deepfake dissemination via provenance tracking,'' in \emph{ACM International Conference on Multimedia (ACMMM)}, 2021, pp. 3546--3555.

\bibitem{yu2021artificial}
N.~Yu, V.~Skripniuk, S.~Abdelnabi, and M.~Fritz, ``Artificial fingerprinting for generative models: Rooting deepfake attribution in training data,'' in \emph{IEEE International Conference on Computer Vision (ICCV)}, 2021, pp. 14\,448--14\,457.

\bibitem{asnani2022proactive}
V.~Asnani, X.~Yin, T.~Hassner, S.~Liu, and X.~Liu, ``Proactive image manipulation detection,'' in \emph{IEEE Conference on Computer Vision and Pattern Recognition (CVPR)}, 2022, pp. 15\,386--15\,395.

\bibitem{yeh2020disrupting}
C.-Y. Yeh, H.-W. Chen, S.-L. Tsai, and S.-D. Wang, ``Disrupting image-translation-based deepfake algorithms with adversarial attacks,'' in \emph{IEEE Winter Conference on Applications of Computer Vision Workshops (WCACV)}, 2020, pp. 53--62.

\bibitem{ruiz2020disrupting}
N.~Ruiz, S.~A. Bargal, and S.~Sclaroff, ``Disrupting deepfakes: Adversarial attacks against conditional image translation networks and facial manipulation systems,'' in \emph{ECCV Workshops}, 2020, pp. 236--251.

\bibitem{segalis2020ogan}
E.~Segalis and E.~Galili, ``Ogan: Disrupting deepfakes with an adversarial attack that survives training,'' \emph{arXiv preprint arXiv:2006.12247}, 2020.

\bibitem{xue2022use}
M.~Xue, C.~Yuan, C.~He, Y.~Wu, Z.~Wu, Y.~Zhang, Z.~Liu, and W.~Liu, ``Use the spear as a shield: An adversarial example based privacy-preserving technique against membership inference attacks,'' \emph{IEEE Transactions on Emerging Topics in Computing(TETC)}, vol.~11, no.~1, pp. 153--169, 2022.

\bibitem{li2020celeb}
Y.~Li, X.~Yang, P.~Sun, H.~Qi, and S.~Lyu, ``Celeb-df: A large-scale challenging dataset for deepfake forensics,'' in \emph{IEEE Conference on Computer Vision and Pattern Recognition (CVPR)}, 2020, pp. 3207--3216.

\bibitem{sun2022faketracer}
P.~Sun, Y.~Li, H.~Qi, and S.~Lyu, ``Faketracer: Exposing deepfakes with training data contamination,'' in \emph{IEEE International Conference on Image Processing (ICIP)}.\hskip 1em plus 0.5em minus 0.4em\relax IEEE, 2022, pp. 1161--1165.

\bibitem{liu2017unsupervised}
M.-Y. Liu, T.~Breuel, and J.~Kautz, ``Unsupervised image-to-image translation networks,'' \emph{Advances in Neural Information Processing Systems (NeurIPS)}, vol.~30, 2017.

\bibitem{karras2018progressive}
T.~Karras, T.~Aila, S.~Laine, and J.~Lehtinen, ``Progressive growing of gans for improved quality, stability, and variation,'' in \emph{International Conference on Learning Representations (ICLR)}, 2018.

\bibitem{karras2019style}
T.~Karras, S.~Laine, and T.~Aila, ``A style-based generator architecture for generative adversarial networks,'' in \emph{IEEE Conference on Computer Vision and Pattern Recognition (CVPR)}, 2019, pp. 4401--4410.

\bibitem{peng2019cgr}
F.~Peng, L.-P. Yin, L.-B. Zhang, and M.~Long, ``Cgr-gan: Cg facial image regeneration for antiforensics based on generative adversarial network,'' \emph{IEEE Transactions on Multimedia (TMM)}, vol.~22, no.~10, pp. 2511--2525, 2019.

\bibitem{karras2020analyzing}
T.~Karras, S.~Laine, M.~Aittala, J.~Hellsten, J.~Lehtinen, and T.~Aila, ``Analyzing and improving the image quality of stylegan,'' in \emph{IEEE Conference on Computer Vision and Pattern Recognition (CVPR)}, 2020, pp. 8110--8119.

\bibitem{agarwal2019protecting}
S.~Agarwal, H.~Farid, Y.~Gu, M.~He, K.~Nagano, and H.~Li, ``Protecting world leaders against deep fakes.'' in \emph{CVPR workshops}, vol.~1, 2019.

\bibitem{li2020identification}
H.~Li, B.~Li, S.~Tan, and J.~Huang, ``Identification of deep network generated images using disparities in color components,'' \emph{Signal Processing}, vol. 174, p. 107616, 2020.

\bibitem{afchar2018mesonet}
D.~Afchar, V.~Nozick, J.~Yamagishi, and I.~Echizen, ``Mesonet: a compact facial video forgery detection network,'' in \emph{2018 IEEE international workshop on information forensics and security (WIFS)}.\hskip 1em plus 0.5em minus 0.4em\relax IEEE, 2018, pp. 1--7.

\bibitem{wang2021representative}
C.~Wang and W.~Deng, ``Representative forgery mining for fake face detection,'' in \emph{IEEE Conference on Computer Vision and Pattern Recognition (CVPR)}, 2021, pp. 14\,923--14\,932.

\bibitem{nataraj2019detecting}
L.~Nataraj, T.~M. Mohammed, B.~Manjunath, S.~Chandrasekaran, A.~Flenner, J.~H. Bappy, and A.~K. Roy-Chowdhury, ``Detecting gan generated fake images using co-occurrence matrices,'' \emph{Electronic Imaging}, vol. 2019, no.~5, pp. 532--1, 2019.

\bibitem{hernandez2020DeepFakeson}
J.~Hernandez-Ortega, R.~Tolosana, J.~Fierrez, and A.~Morales, ``Deepfakeson-phys: Deepfakes detection based on heart rate estimation,'' \emph{arXiv preprint arXiv:2010.00400}, 2020.

\bibitem{ciftci2020fakecatcher}
U.~A. Ciftci, I.~Demir, and L.~Yin, ``Fakecatcher: Detection of synthetic portrait videos using biological signals,'' \emph{IEEE Transactions on Pattern Analysis and Machine Intelligence (TPAMI)}, 2020.

\bibitem{zhao2021learning}
T.~Zhao, X.~Xu, M.~Xu, H.~Ding, Y.~Xiong, and W.~Xia, ``Learning self-consistency for deepfake detection,'' in \emph{IEEE International Conference on Computer Vision (ICCV)}, 2021, pp. 15\,023--15\,033.

\bibitem{zhang2018unreasonable}
R.~Zhang, P.~Isola, A.~A. Efros, E.~Shechtman, and O.~Wang, ``The unreasonable effectiveness of deep features as a perceptual metric,'' in \emph{IEEE Conference on Computer Vision and Pattern Recognition (CVPR)}, 2018, pp. 586--595.

\bibitem{wang2020cnn}
S.-Y. Wang, O.~Wang, R.~Zhang, A.~Owens, and A.~A. Efros, ``Cnn-generated images are surprisingly easy to spot... for now,'' in \emph{IEEE Conference on Computer Vision and Pattern Recognition (CVPR)}, 2020, pp. 8695--8704.

\end{thebibliography}

 
\vspace{-1.0cm}
\begin{IEEEbiography}[{\includegraphics[width=1.0in,height=1.25in,clip,keepaspectratio]{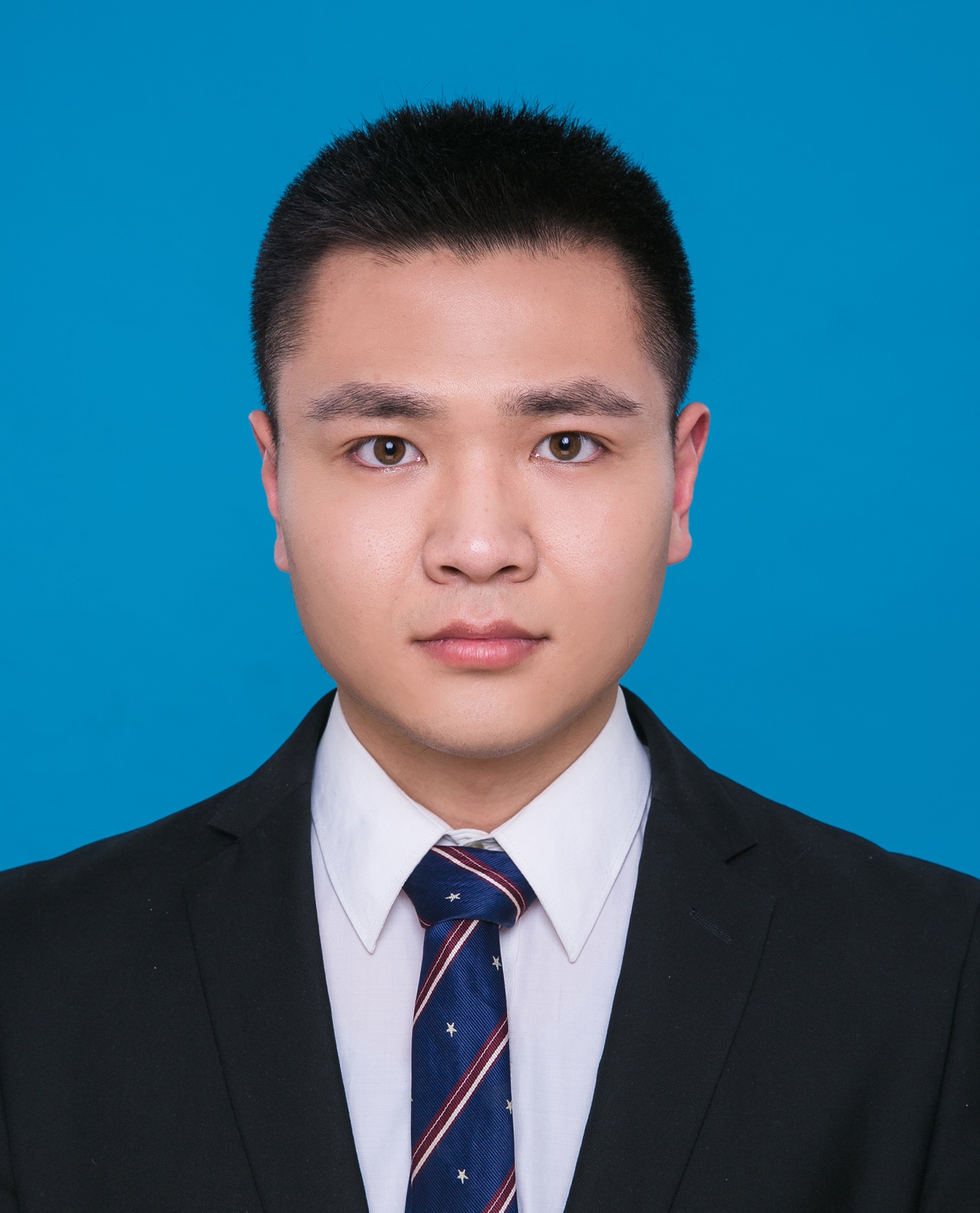}}]{Pu Sun} is a Ph.D. candidate in the School of Computer Science and Technology, at University of Chinese Academy of Sciences. He received M.S. degree in computer technology at University of Chinese Academy of Sciences in 2021 and B.S. degree in Information Engineering in 2018 at China University of Mining and Technology. His research interest is mainly focused on artificial intelligence security and multimedia forensics.
\end{IEEEbiography}
\vspace{-1.0cm}
\begin{IEEEbiography}[{\includegraphics[width=1.0in,height=1.25in,clip,keepaspectratio]{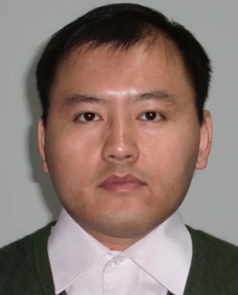}}]{Honggang Qi} (Member, IEEE) received the M.S. degree in computer science from Northeast University, Shenyang, China, in 2002, and the Ph.D. degree in computer science from the Institute of Computing Technology, Chinese Academy of Sciences, Beijing, China, in 2008. 
He is currently a Professor with the School of Computer Science and Technology, University of Chinese Academy of Sciences. His current research interests include computer vision, video coding, and very large scale integration design.
\end{IEEEbiography}
\vspace{-1.0cm}
\begin{IEEEbiography}[{\includegraphics[width=1.0in,height=1.25in,clip,keepaspectratio]{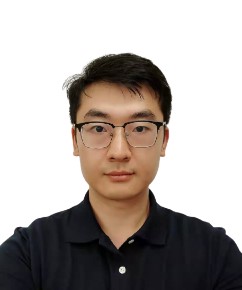}}]{Yuezun Li} is a lecturer in the Center on Artificial Intelligence, at Ocean University of China. He was a Senior Research Scientist at the Department of Computer Science and Engineering of the University at Buffalo, SUNY. He received Ph.D. degree in computer science at University at Albany, SUNY in 2020. He received M.S. degree in Computer Science in 2015 and B.S. degree in Software Engineering in 2012 at Shandong University. Dr. Li’s research interest is mainly focused on artificial intelligence security and multimedia forensics. His work has been published in peer-reviewed conferences and journals, including ICCV, CVPR, TCSVT, TNNL, etc.
\end{IEEEbiography}

\vspace{-1.0cm}

\begin{IEEEbiography}[{\includegraphics[width=1.0in,height=1.25in,clip,keepaspectratio]{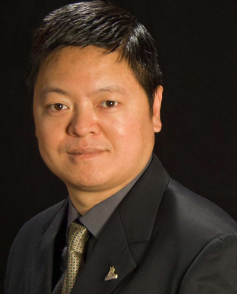}}]{Siwei Lyu} (Fellow, IEEE) received the B.S. degree in information science and the M.S. degree in computer science from Peking University, China, in 1997 and 2000, respectively, and the Ph.D. degree in computer science from Dartmouth College in 2005. From 2005 to 2008, he was a Post-Doctoral Research Associate with Howard Hughes Medical Institute and the Center for Neural Science of New York University. He is currently a Full Professor in computer science with the University at Buffalo (State University of New York at Buffalo). His research interests include digital media forensics, computer vision, and machine learning.
\end{IEEEbiography}

\vfill

\end{document}